\documentclass[lettersize,journal]{IEEEtran}
\usepackage[utf8]{inputenc}
\usepackage{bm}
\usepackage{amsmath}
\newcommand\norm[1]{\left\lVert#1\right\rVert}
\usepackage[pdftex]{graphicx}
 \usepackage{algpseudocode}
\usepackage{algorithm}
\usepackage{mathtools}
\usepackage{amsmath}
\usepackage{amssymb}
\usepackage{xcolor}

\usepackage{amsthm}
\usepackage{enumitem}
  \newtheoremstyle{dotless}{}{}{\itshape}{}{\bfseries}{}{ }{}
  \theoremstyle{dotless}
  \newtheorem{thm}{Theorem}
  \newtheorem{prop}{Proposition}

\begin{document}

\title{Unknown Face Presentation Attack Detection via Localised Learning of Multiple Kernels}

\author{Shervin Rahimzadeh Arashloo~\IEEEmembership{}
\thanks{S.R. Arashloo is with the centre for vision, speech, and signal processing (CVSSP), University of Surrey, Guildford, Surrey, UK. e-mail: s.rahimzadeh@surrey.ac.uk}
\thanks{Manuscript received ? ?, 2022; revised ? ?, 2022.}}

\markboth{}%
{}

\IEEEpubid{}

\maketitle

\begin{abstract}
The paper studies face spoofing, a.k.a. presentation attack detection (PAD) in the demanding scenarios of unknown types of attack. While earlier studies have revealed the benefits of ensemble methods, and in particular, a multiple kernel learning approach to the problem, one limitation of such techniques is that they typically treat the entire observation space similarly and ignore any variability and \textit{local} structure inherent to the data. This work studies this aspect of the face presentation attack detection problem in relation to multiple kernel learning in a one-class setting to benefit from intrinsic local structure in bona fide face samples. More concretely, inspired by the success of the one-class Fisher null formalism, we formulate a convex \textit{localised} multiple kernel learning algorithm by imposing a joint matrix-norm constraint on the collection of local kernel weights and infer locally adaptive weights for zero-shot one-class unseen attack detection.

We present a theoretical study of the proposed localised MKL algorithm using Rademacher complexities to characterise its generalisation capability and demonstrate the advantages of the proposed technique over some other options. An assessment of the proposed approach on general object image datasets illustrates its efficacy for abnormality and novelty detection while the results of the experiments on face PAD datasets verifies its potential in detecting unknown/unseen face presentation attacks.
\end{abstract}


\section{Introduction}
When exposed to sensor-level attacks, known as presentation/spoofing attacks, the trustworthiness of face recognition systems \cite{8063938,8713930,8752382,9184823} is significantly questioned, necessitating the development of robust presentation attack detection algorithms. Despite big advances, the current face PAD approaches have their own limitations. One source of difficulty in detecting presentation attacks is the diversity in attack samples which may be attributed to using a variety of different presentation attack mechanisms such as print, replay, mask, make-up, or to other environmental imaging conditions resulting in a significant within-class variance that may erode the separability between classes. Unseen/unknown attacks that were not accessible during the system's training phase represent another, arguably a severe challenge in face PAD, rendering it as an open-set classification problem \cite{6365193}. In this respect, when the test-phase attacks are fundamentally different from those observed during the training phase, the performance of the classical two-class closed-set systems may be seriously compromised. Due to its significant practical importance, detection of unseen presentation attacks has received considerable attention both in face \cite{9303455,7984788,9535490,8411206,8698574,8682253,8953776,DAD} as well as in other biometric modalities \cite{6996254,7180344,7823572,8897269,Sahidullah2019}. 

Although other alternatives may exist, the one-class classification methodology \cite{khan_madden_2014,Tax2004} presents a promising strategy to the unknown face presentation attack detection problem. Specifically, the undesired training bias towards seen attacks may be removed by utilising only genuine samples for one-class training and solely trying to capture the essence of the bona fide observations. Various studies have looked into the functionality of this approach to detect unknown/unseen face presentation attacks \cite{7984788,9303455,FATEMIFAR2021107696} where it has been observed that a pure one-class learning approach trained solely on bona fide samples may yield better generalisation capability against unseen attacks compared to the closed-set two-class formulations. Although single-learner one-class face PAD systems have been developed \cite{8698574,FATEMIFAR2021107696,8682253,DBLP:journals/corr/abs-1904-02860}, currently, their capabilities may be insufficient to meet the stringent requirements of a reliable and secure real-world system.

Among others, ensemble learning serves as a well-known strategy for improving pattern classification system performance, in which several knowledge sources for a pattern are merged for the final decision. The variation of this generic methodology can be related to the level at which the combination is realised, such as data-level, feature-level, soft-, or hard-decision level, etc. In the case of face PAD, this approach has been effectively implemented in a pure one-class setting using a classifier combination \cite{8987326,9190814,9506251,FATEMIFAR2022} or a multiple kernel fusion \cite{9303455,9535490} scheme with the ensemble system being observed to outperform each individual learner/representation enrolled in the ensemble. In particular, resorting to a reproducing kernel Hilbert space (RKHS) \cite{Cortes1995} fusion facilitates benefiting from learning strategies with a strong basis in statistical learning theory \cite{Vapnik1998,JMLR:v12:kloft11a} while allowing the learning of the classifier to be separated from the representations. Furthermore, by constructing each individual kernel based on deep representations, a multiple kernel system can benefit from deep convolutional representations for improved detection \cite{9535490,9303455}. As an instance of the kernel-based ensemble methods, the approach in \cite{9535490}, seeks to infer one-class kernel weights to be fused for a Fisher null-space classifier in the presence of a matrix-norm regularisation on kernel weights. An experimental evaluation of the method led to promising results for detecting unseen attacks. In spite of the relative success of the one-class ensemble-based techniques for face PAD, one limitation of the existing methods is that they ignore any inherent variation in the data and treat the entire observation space similarity by associating global weights to each representation/learner irrespective of the local region where the pattern resides. This is an unrealistic assumption since, compared to other representations/learners, each particular representation/learner may be more effective in certain regions of the observation space while providing less discriminative capability in others. As such, a local strategy that considers potential variability in genuine observations may be advantageous. This aspect of the problem is overlooked in the existing ensemble one-class face PAD methods.

In the context of multiple kernel learning (MKL), while the vast majority of existing algorithms are global in nature, meaning the kernel weights are shared across the whole observation space, the issue was initially identified in \cite{DBLP:conf/icml/GonenA08} and has received considerable attention since then \cite{5459172,MU20111095,6129537,7792117,7896638}. However, existing localised multiple kernel learning methods use non-convex objective functions which raises questions with regards to their generalisation capacities. Another potential drawback of a non-convex formulation is the computational difficulty in obtaining a good local optima, as well as the ensued hurdles in reproducibility due to variable local optima. Furthermore, the majority of existing local MKL techniques' theoretical performance guarantees are poorly understood. An exception to these is the study in \cite{pmlr-v63-lei63} which formulates the MKL problem as a convex optimisation task. Yet, certain limitations apply to this method too. First, the MKL problem is formulated in the presence of an $\ell_p$-norm constraint on kernel weights. More recent studies have demonstrated that a more general matrix $\ell_{p,q}$-norm regularisation on kernel weights may yield better performance in different applications settings \cite{9535490,NIPS2012_fccb3cdc,7792117,8259375}. Second, the localised MKL method in \cite{pmlr-v63-lei63} decouples the learning tasks into a set of disjoint MKL problems each associated with a distinct region in the data. Such an approach completely ignores any dependencies which might exist between kernel weights corresponding to different clusters. Third, the work in \cite{pmlr-v63-lei63} focuses on a multi-/binary-class SVM-based objective function and does not provide a straightforward extension for the one-class Fisher null-space classification, which has demonstrated superior performance for the unknown face spoofing detection problem.

In this paper, we study the locality information for unknown face PAD in the context of ensemble techniques and try to capture and use the inherent local structure in bona fide samples for improved detection performance. More specifically, drawing on the one-class MKL method of \cite{9535490}, we formulate a \textit{local} multiple kernel learning one-class method addressing the aforementioned limitations of the existing localised MKL approaches and illustrate the utility of the proposed technique in detecting unknown/unseen face presentation attacks. In particular, in the proposed method, we regularise kernel weights using a \textit{joint} matrix-norm constraint as compared with the existing local MKL methods using disjoint vector $\ell_p$-norm regularisation. By imposing a \textit{joint} norm constraint on the collection of kernel weights from all the existing clusters one enables an active interaction between kernel weights corresponding to different clusters which is in sharp contrast with the earlier studies where disjoint regularisation constraints were considered in the localised methods. Mathematically, we illustrate that incorporating locality into the MKL framework amounts to increasing the number of effective kernels (by a factor equal to the number of clusters) in the ensemble which provides further explanation for its superior performance as compared with its non-localised variant. We theoretically study the proposed approach and illustrate its merits in terms of the empirical Rademacher complexities compared to some other alternatives. Experimentally, we evaluate the proposed localised one-class MKL algorithm on different datasets and provide a comparison to the state-of-the-art techniques.

\subsection{Main contributions}
A summary of the major contributions of this work is as follows.
\begin{itemize}
    \item We propose a \textit{localised} multiple kernel learning method for pure one-class classification of unknown/unseen face presentation attacks. By virtue of a localised formulation, the proposed approach is able to capture and benefit from the intrinsic variability and local structure inherent to bona fide samples for improved detection performance;
    \item We formulate the learning problem of the proposed localised MKL approach as a convex optimisation problem and present an effective method to solve it;
    \item A theoretical study of the generalisation performance of the proposed method based on Rademacher complexities is conducted and a comparison against other alternatives is provided;
    \item We evaluate the proposed localised MKL one-class approach for general object image abnormality and novelty detection as well as unseen face PAD and compare it against state-of-the-art approaches from the literature.
\end{itemize}

The rest of the paper is structured as detailed next. In Section \ref{RW}, we review the relevant literature on face PAD with a particular attention to the unknown/unseen and ensemble face presentation attack detection methods. Section \ref{PR} briefly reviews the Fisher null projection for one-class classification and its extension to an MKL setting as presented in \cite{9535490}. Section \ref{PRO} presents a new \textit{localised} one-class Fisher null approach and then extends it to a multiple kernel learning setting followed by an optimisation approach to solve the corresponding convex optimisation task. Section \ref{TA} theoretically studies the proposed approach in terms of Rademacher complexities to derive a generalisation error bound for the proposed method and theoretically compares the proposed algorithm against some other alternatives. In Section \ref{ER}, we present the results of evaluating the proposed approach on multiple datasets. Finally, Section \ref{conc} brings the paper to a conclusion.

\section{Prior Work}
\label{RW}
Challenge-response, hardware-, and software-based strategies \cite{edmunds:tel-01576830, Bhattacharjee2019,Ramachandra:2017:PAD:3058791.3038924,DBLP:series/acvpr/978-3-319-92626-1} have all been put into practice to address the face PAD problem, with software-based alternatives gaining the most traction. Software-based approaches detect a presentation attack in a video/image by utilising various intrinsic features of signal content. This work uses a face PAD software formalism using RGB photos. Regarding cues employed in the software-based category, texture \cite{BOULKENAFET20181,Peng2018} may be considered as the most extensively deployed cue for presentation attack detection, but there are other motion-based methods \cite{KOLLREIDER2009233,FENG2016451,7185398}. Some other studies \cite{6976921,7351280,9294085} employ shape or colour information or spoofing medium contours \cite{8884098} for face spoofing detection, while by using frequency information content in the Fourier domain, frequency-based approaches \cite{6199760,7017526,7185398, 6612957} try to detect presentation attacks. Other study \cite{8920060} have employed a statistical method to model noise in the facial PAD while deep convolutional neural networks serve as a well-known tool for detecting presentation attacks \cite{DBLP:journals/corr/abs-1806-07492,DBLP:journals/corr/YangLL14,8714076}. 

From a classification viewpoint, the classical method to detect face presentation attacks is the closed-set two-class formulation where both genuine and attack samples are utilised to train the system. Linear discriminant analysis \cite{6976921,HAMDAN201875}, SVMs \cite{Heusch2019,LI2018182}, neural networks \cite{FENG2016451} and CNNs \cite{DBLP:journals/corr/abs-1806-07492,DBLP:journals/corr/YangLL14}, Bayesian networks \cite{EDMUNDS2018314} and Adaboost techniques \cite{7880281}, constitute some of the most extensively used two-class methods. Regression-based procedures \cite{7041231}, in contrast, attempt to map the input features to the associated labels.

One-class classification provides a promising alternative to the classical two-class formulation for face presentation attack detection and has been found to be especially useful in the case of unseen attacks \cite{7984788}. As an instance, the work in \cite{8411206} employs a Gaussian mixture model to estimate the distribution of bona fide samples using quality metrics of images. Auto-Encoder and One-Class SVM are used to detect unknown attacks in \cite{8698574} while \cite{FATEMIFAR2021107696,8682253} propose a subject-specific modelling strategy to construct separate one-class classifiers for each person. Other work \cite{DBLP:journals/corr/abs-1904-02860} considers the unseen zero-shot face PAD using a tree network while \cite{DAD} uses a triplet focal objective function for metric learning.

Ensemble methods, in which the final decision for an object is generated as a consensus of multiple learners, constitute a relatively successful subclass of one-class algorithms for detecting unknown face presentation attacks. As an instance of the work in this group, the study in \cite{8987326} considers a weighted linear combination of normalised classifier output scores where the combination weights are inferred using both negative and positive observations. However, such a weight adjusting technique raises issues about the generalisability of the learned weights to unknown attacks. The work in \cite{9190814}, considers an ensemble of one-class learners fused via a stacking method and presents a two-stage Genetic Algorithm to improve the generalisation error by using only genuine samples for training. In a subsequent study \cite{9506251}, the authors propose to optimise the parameters of a weighted averaging of multiple classifiers through a two-stage optimisation approach composed of Particle Swarm and Pattern Search algorithms to minimise the possibility of local minimum in weight learning. A new classifier score normalisation method for heavy-tailed distributions is presented to facilitate effective combination of multiple learners. Evaluation of the algorithm on multiple datasets illustrated promising performance of this method. However, since the fusion mechanism in a classifier combination scheme is based solely on the scores provided by individual classifiers, any extra information that could be potentially beneficial is omitted. Another study \cite{9303455} considers a multiple kernel fusion approach, revealing that a kernel fusion strategy can lead to substantial improvements in the performance in comparison to each base kernel. Unlike the classifier fusion paradigm, in a kernel combination method, the information content related to each representation is not summarised merely in terms of a single score, allowing the ensemble system to access a more informative representation of the data. The relative discriminative potential of each kernel, however, is not taken into account in \cite{9303455}, and all representations in the composite kernel are weighted equally. Other study \cite{9535490} considers a multiple kernel learning algorithm that could capture the inherent sparsity within a problem and take into account potential discrepancies in the discriminative content of different kernels and infer kernel weights in a one-class learning setting to address this shortcoming. The study in \cite{9535490} presented a number of desired attributes to build an ensemble face PAD system. First, instead of classifier scores, it operates on kernel-space representations, and thus, could access a richer information content for classification. Second, each representation is weighted in the composite kernel where the weights are learned using genuine samples only. Third, the multiple kernel learning approach could minimise the contribution of a potentially less useful representation to the compound kernel, thanks to deploying a matrix-norm regularisation constraint inducing sparsity. And last but not the least, it is formulated as a convex optimisation problem, facilitating optimisation.

Despite their advantages with respect to single-learner PAD systems, the existing one-class ensemble face PAD techniques have the limitation that they ignore any inherent variation in the bona fide data and treat the entire observation space similarly by assigning a global weight to each representation/learner, regardless of the locale where the model operates. This is quite a restrictive assumption since each representation/learner may be more effective in some regions of the observation space while providing less discriminative capability in others. A \textit{local} learning strategy may thus be helpful as it takes into account potential variation in genuine observations, which is generally neglected in the existing ensemble one-class face PAD methods. This work addresses this aspect of the ensemble one-class face presentation attack detection in the context of multiple kernel learning. By virtue of using local structure, the proposed approach assigns locally adaptive weights to different representations in distinct regions of the reproducing kernel Hilbert space and learns the relative importance of the incorporated representations using a convex formulation.

\section{Preliminaries}
\label{PR}
The learning problem for the regularised null-space one-class kernel Fisher method in the primal space is \cite{9055448}
\begin{eqnarray}
\min_{\boldsymbol\eta} \frac{\theta}{n}\sum_{i=1}^n\big(1-\boldsymbol\psi(\mathbf{x}_i)^\top\boldsymbol\eta\big)^2+\norm{\boldsymbol\eta}_2^2
\label{fn}
\end{eqnarray}
\noindent where $\mathbf{x}_i|_{i=1}^n$ denotes the set of positive training observations and $\boldsymbol\psi(\mathbf{x}_i)$ stands for the feature associated with $\mathbf{x}_i$ in the Hilbert space.
The optimal solution to the problem above yields a linear model in the Hilbert space, i.e. $f(.)=\boldsymbol\psi(.)^\top\boldsymbol\eta$, that maps observations onto the regularised Fisher null space for classification. In particular, positive samples are encouraged to be mapped closer to point $"1"$ while non-target test observations are expected to be projected further away. In practice, however, it is more convenient to consider the dual of the problem in Eq. \ref{fn} given as \cite{9535490,9055448}:
\begin{eqnarray}
\max_{\boldsymbol{\lambda}} -\boldsymbol{\lambda}^\top\mathbf{K}\boldsymbol{\lambda}-\delta \boldsymbol{\lambda}^\top \boldsymbol{\lambda}+2\boldsymbol{\lambda}^\top\mathbf{1}
\label{dual}
\end{eqnarray}
where $\mathbf{K}$ stands for the kernel matrix, the element of which at the $j^{th}$ column and $i^{th}$ row is given as $\mathbf{K}^{i,j}=\boldsymbol\psi(\mathbf{x}_i)^\top\boldsymbol\psi(\mathbf{x}_j)=\kappa(\mathbf{x}_i,\mathbf{x}_j)$ and $\kappa(.,.)$ stands for the kernel and $\delta=n/\theta$. The optimal solution to Eq. \ref{dual} may be determined by setting the gradient of the objective function w.r.t. $\boldsymbol\lambda$ to zero to obtain $\boldsymbol\lambda=(\mathbf{K}+\delta\mathbf{I})^{-1}\mathbf{1}$, where $\mathbf{I}$ stands for the identity matrix, and $\mathbf{1}$ is a vector of $1$'s of dimension $n$. Once the optimal $\boldsymbol\lambda$ is computed, an observation $\mathbf{y}$ is mapped onto the Fisher null-space as
\begin{eqnarray}
f(\mathbf{y})=\Big[\kappa(\mathbf{y},\mathbf{x}_1),\dots,\kappa(\mathbf{y},\mathbf{x}_n)\Big]\boldsymbol\lambda
\end{eqnarray}

The work in \cite{9535490} extends the one-class kernel Fisher null method to an MKL setting by imposing a matrix $\ell_{p,q}$-norm constraint ($p,q\geq1$) on kernel weights. Assuming $G$ base kernels $\mathbf{K}_g$, $g = 1,\dots,G$ which are to be fused linearly using the non-negative weights $\mu_g$'s, the MKL problem considered in \cite{9535490} is
\begin{eqnarray}
\nonumber \min_{\boldsymbol\mu}\max_{\boldsymbol\lambda} &-\boldsymbol{\lambda}^\top\big(\sum_{g=1}^G\mu_{g}\mathbf{K}_{g}\big)\boldsymbol{\lambda}-\delta \boldsymbol{\lambda}^\top \boldsymbol{\lambda}+2\boldsymbol{\lambda}^\top\mathbf{1}\\
&\text{s.t. } \boldsymbol\mu\geq 0, \norm{\boldsymbol\mu\boldsymbol\mu^\top}_{p,q}\leq 1
\label{lpq}
\end{eqnarray}
where the non-negativity constraint on kernel weights $\mu_g$'s, collectively represented as vector $\boldsymbol\mu$, ensures that the combined kernel matrix is valid while $\norm{.}_{p,q}$ denotes a matrix $\ell_{p,q}$-norm ($p,q\geq1$) defined as
\begin{eqnarray}
\norm{\boldsymbol{\Upsilon}}_{p,q} = \Bigg(\sum_{s} \Big(\sum_{t}|\upsilon_{ts}|^p\Big)^{q/p}\Bigg)^{1/q}
\end{eqnarray}
\noindent where $s$ and $t$ index the columns and the rows of the matrix, respectively, and $\upsilon_{ts}$ stands for the element at the $t^{th}$ row and $s^{th}$ column of matrix $\boldsymbol\Upsilon$. A matrix mixed-norm constraint on kernel weights not only provides a controlling mechanism over sparsity of kernel weights but also enables interactions between kernels \cite{8259375}. In contrary, the more commonly used vector $\ell_p$-norm regularisation \cite{JMLR:v12:kloft11a} lacks an explicit mechanism to capture inter-kernel interactions. As such, the matrix-norm regularisation has been observed to outperform its vector-norm counterpart in different settings \cite{9535490,NIPS2012_fccb3cdc,7792117,8259375}. Note that the vector-norm constraint may be considered as a special case of the matrix-norm constraint as setting $p=q$ would reduce the matrix-norm into a vector-norm.

In \cite{9535490}, for solving Eq. \ref{lpq}, $\mathbf{u}$ is assumed as a $G$-dimensional vector the $g^{th}$ element of which is defined as $u_g \coloneqq \boldsymbol\lambda^\top\mathbf{K}_g\boldsymbol\lambda$. By applying the minimax theorem \cite{VonNeumann1971,DPa95} and switching the order of optimisation, the problem in Eq. \ref{lpq} is expressed as
\begin{eqnarray}
\nonumber \max_{\boldsymbol{\lambda}}\Big\{-\delta \boldsymbol{\lambda}^\top \boldsymbol{\lambda}+2\boldsymbol{\lambda}^\top\mathbf{1}+\min_{\boldsymbol{\mu}}-\boldsymbol{\mu}^\top\mathbf{u}\Big\}\\
\text{s.t. }
\boldsymbol\mu \geq 0, \norm{\boldsymbol{\mu}\boldsymbol\mu^\top}_{p,q}\leq 1
\label{pi1}
\end{eqnarray}
Next, an iterative two-step method based on an analytical solution in kernel weights is proposed in \cite{9535490} to determine the optimisers of the saddle point MKL problem in Eq. \ref{pi1}.

\section{The proposed method}
\label{PRO}
In this section, first, we present a novel \textit{localised} variant of the regularised null-space Fisher null approach, and then extend the proposed localised method to a multiple kernel learning setting.
\subsection{Localised Fisher null projection}
In order to incorporate localised information into the one-class kernel Fisher null method, let us assume that the target/positive training set $\mathbf{x}_i|_{i=1}^n$ is probabilistically grouped into $C$ clusters where $p_c(\mathbf{x}_i)$ denotes the probability of observation $\mathbf{x}_i$ being assigned to the cluster indexed by $c$. Inspired by \cite{pmlr-v63-lei63}, we consider a convex combination of $C$ local models each associated with a different cluster and use cluster likelihoods to form the projection function. More explicitly, the projection function in the localised model is defined as
\begin{eqnarray}
f(\mathbf{x}_i)=\sum_{c=1}^Cp_c(\mathbf{x}_i)f_c(\mathbf{x}_i)
\label{llocal}
\end{eqnarray}
where each local model $f_c(.)$ is characterised as $f_c(\mathbf{x}_i)=\boldsymbol\psi(\mathbf{x}_i)^\top\boldsymbol\eta_c$. 
Considering $\boldsymbol\psi_c(\mathbf{x}_i)\coloneqq p_c(\mathbf{x}_i)\boldsymbol\psi(\mathbf{x}_i)$, the primal problem for the proposed localised one-class kernel Fisher null approach then reads
\begin{eqnarray}
\min_{\boldsymbol\eta_c's} \frac{\theta}{n}\sum_{i=1}^n\big[1-\sum_{c=1}^C\boldsymbol\psi_c(\mathbf{x}_i)^\top\boldsymbol\eta_c\big]^2+\sum_{c=1}^C\norm{\boldsymbol\eta_c}_2^2
\label{pl}
\end{eqnarray}
where the first term captures the empirical loss while the second term imposes a Tikhonov regularisation on the parameters of the localised model.
\subsection{Dualisation}
As it is common in kernel-based methods, we consider the dual of the problem in Eq. \ref{pl}. For this purpose, the problem in Eq. \ref{pl} can be expressed as
\begin{eqnarray}
\nonumber \min_{\boldsymbol\eta_c's}&\frac{1}{\delta}\norm{\boldsymbol\tau}_2^2+\sum_{c=1}^C\norm{\boldsymbol\eta_c}_2^2\\
&\text{s.t. }\boldsymbol\tau=\mathbf{1}-\sum_{c=1}^C\boldsymbol\Psi_c(\mathbf{X})^\top\boldsymbol\eta_c
\end{eqnarray}
\noindent where $\boldsymbol\Psi_c(\mathbf{X})$ denotes a matrix collection of $\boldsymbol\psi_c(\mathbf{x}_i)$'s for $\mathbf{x}_i|_{i=1}^n$, $\mathbf{1}$ stands for an $n$-vector of one's and $\delta=n/\theta$. The Lagrangian of the constrained problem above is
\begin{eqnarray}
\mathcal{L} = \frac{1}{\delta}\norm{\boldsymbol\tau}_2^2+\sum_{c=1}^C\boldsymbol\eta_c^\top\boldsymbol\eta_c+\boldsymbol\omega^\top(\mathbf{1}-\sum_{c=1}^C\boldsymbol\Psi_c(\mathbf{X})^\top\boldsymbol\eta_c-\boldsymbol\tau)
\label{lag1}
\end{eqnarray}
where $\boldsymbol\omega$ denotes the vector of Lagrange multipliers. Next, one has to minimise $\mathcal{L}$ in $\boldsymbol\eta_c$'s and $\boldsymbol\tau$ and then maximise it w.r.t. $\boldsymbol\omega$. For the minimisation of $\mathcal{L}$ w.r.t. $\boldsymbol\eta_c$'s and $\boldsymbol\tau$, the derivatives with respect to these variables are set to zero. Omitting the intermediate steps, the Lagrangian is obtained as
\begin{eqnarray}
\mathcal{L} = \frac{-1}{4}\boldsymbol\omega^\top\big(\sum_{c=1}^C\mathbf{K}_c\big)\boldsymbol\omega-\frac{\delta}{4}\boldsymbol\omega^\top\boldsymbol\omega+\boldsymbol\omega^\top\mathbf{1}
\end{eqnarray}
\noindent where $\mathbf{K}_c$ denotes the kernel matrix associated with cluster $c$ whose $(i,j)^{th}$ element is $\mathbf{K}_c^{i,j}\coloneqq\boldsymbol\psi_c(\mathbf{x}_i)^\top\boldsymbol\psi_c(\mathbf{x}_j)=p_c(\mathbf{x}_i)\kappa(\mathbf{x}_i,\mathbf{x}_j)p_c(\mathbf{x}_j)$. Applying a variable change as $\boldsymbol\lambda = \boldsymbol\omega/2$, the dual problem for the proposed localised one-class null-space Fisher approach is expressed as
\begin{eqnarray}
\max_{\boldsymbol{\lambda}} -\boldsymbol{\lambda}^\top\big(\sum_{c=1}^C\mathbf{K}_c\big)\boldsymbol{\lambda}-\delta \boldsymbol{\lambda}^\top \boldsymbol{\lambda}+2\boldsymbol{\lambda}^\top\mathbf{1}
\label{ldual}
\end{eqnarray}
Setting the gradient of the function above w.r.t $\boldsymbol\lambda$ to zero yields the discriminant in the Hilbert space as $\boldsymbol\lambda=\big(\sum_{c=1}^C\mathbf{K}_c\big)^{-1}\mathbf{1}$. After computing $\boldsymbol\lambda$, an observation $\mathbf{y}$ would be mapped onto the localised Fisher null-space as
\begin{eqnarray}
f(\mathbf{y})=\sum_{c=1}^Cp_c(\mathbf{y})\Big[\kappa(\mathbf{y},\mathbf{x}_1)p_c(\mathbf{x}_1),\dots,\kappa(\mathbf{y},\mathbf{x}_n)p_c(\mathbf{x}_n)\Big]\boldsymbol\lambda
\end{eqnarray}

As may be observed from Eq. \ref{ldual}, the optimisation problem associated with the localised approach resembles that of its non-localised counterpart (see Eq. \ref{dual}) except for the fact that the effective kernel matrix in the localised model corresponds to a sum of local kernel matrices. As such, the extensions and optimisation procedures corresponding to the non-localised approach may be adapted and applied to the new localised model. In this context, of particular interest to this study is a multiple kernel learning extension, discussed next.

\subsection{Multiple kernel localised Fisher null projection}
For the extension of the proposed one-class localised Fisher null approach to a multiple kernel learning setting, let us assume there exist $G$ base kernels, $\kappa_g(.,.)$, $g=1, \dots, G$ which are to be combined linearly using non-negative weights. Inspired by the success of a mixed matrix-norm constraint for multiple kernel one-class classification \cite{9535490}, in this study, we consider an $\ell_{p,q}$-norm regularisation ($p,q\geq1$) for kernel weights. By replacing the kernel matrix of each cluster in Eq. \ref{ldual} with a weighted linear fusion of $G$ kernels and imposing matrix-norm and non-negativity constraints on kernel weights, the learning problem associated with the proposed localised $\ell_{p,q}$-norm one-class MKL method is defined as
\begin{eqnarray}
\nonumber \min_{\boldsymbol\mu}\max_{\boldsymbol\lambda} &-\boldsymbol{\lambda}^\top\big(\sum_{c=1}^C\sum_{g=1}^G\mu_{cg}\mathbf{K}_{cg}\big)\boldsymbol{\lambda}-\delta \boldsymbol{\lambda}^\top \boldsymbol{\lambda}+2\boldsymbol{\lambda}^\top\mathbf{1}\\
&\text{s.t. } \boldsymbol\mu\geq 0, \norm{\boldsymbol\mu\boldsymbol\mu^\top}_{p,q}\leq 1
\label{lmkl}
\end{eqnarray}
\noindent where $\mathbf{K}_{cg}$ denotes the kernel matrix associated with the kernel function $\kappa_g(.,.)$ for the cluster indexed by $c$. That is, the $(i,j)^{th}$ element of matrix $\mathbf{K}_{cg}$ is $\mathbf{K}_{cg}^{i,j} \coloneqq p_c(\mathbf{x}_i)\kappa_g(\mathbf{x}_i,\mathbf{x}_j)p_c(\mathbf{x}_j)$. In Eq. \ref{lmkl}, $\boldsymbol\mu$ is defined as a $CG$-dimensional vector formed as the collection of kernel weights from all clusters, i.e. $\boldsymbol\mu=[\boldsymbol\mu_1^\top,\dots,\boldsymbol\mu_C^\top]^\top$ where $\boldsymbol\mu_c=[\mu_{c1},\dots,\mu_{cG}]^\top$ is the kernel weight vector associated with cluster $c$. Since $\boldsymbol\mu$ is formed as a collection of kernel weights from all clusters, a norm constraint imposed on $\boldsymbol\mu$ \textit{jointly} affects the kernel weights corresponding to all clusters, and thus, promotes interdependence between cluster weights. This is in contrast to a case where one may impose \textit{disjoint} constraints on the kernel weights associated with different clusters \cite{pmlr-v63-lei63} which would lack an explicit interaction mechanism between cluster kernel weights once the data is clustered. A theoretical analysis regarding the generalisation performance of the proposed joint norm constraint over all clusters and its advantages compared to the disjoint case is presented in Section \ref{TA}.
\subsection{Optimisation}
\label{opt}
For the optimisation of the saddle point problem in Eq. \ref{lmkl}, let us define the $CG$-element vector $\mathbf{u}$ as $\mathbf{u}\coloneqq[\mathbf{u}_1^\top,\mathbf{u}_2^\top,\dots,\mathbf{u}_C^\top]^\top$ where the $G$-dimensional vector $\mathbf{u}_c$ is
\begin{eqnarray}
\mathbf{u}_c=[\boldsymbol\lambda^\top\mathbf{K}_{c1}\boldsymbol\lambda,\boldsymbol\lambda^\top\mathbf{K}_{c2}\boldsymbol\lambda,\dots,\boldsymbol\lambda^\top\mathbf{K}_{cG}\boldsymbol\lambda]
\end{eqnarray}
Eq. \ref{lmkl} may now be expressed as
\begin{eqnarray}
\nonumber \min_{\boldsymbol\mu}\max_{\boldsymbol\lambda} &-\delta\boldsymbol\lambda^\top\boldsymbol\lambda+2\boldsymbol\lambda^\top\mathbf{1}-\boldsymbol\mu^\top\mathbf{u}\\
&\text{s.t. } \boldsymbol\mu\geq 0, \norm{\boldsymbol\mu\boldsymbol\mu^\top}_{p,q}\leq 1
\label{fobj1}
\end{eqnarray}
Note that for fixed $\boldsymbol\lambda$, the optimisation problem in Eq. \ref{fobj1} is convex w.r.t. $\boldsymbol\mu$ since the constraints imposed by the matrix-norm may be characterised as the supremum of a set of linear functions of $\boldsymbol\mu\boldsymbol\mu^\top$, and consequently, give rise to a set which is convex \cite{8259375,BV2014}. Furthermore, when $\boldsymbol\mu$ is fixed, the loss is concave w.r.t. $\boldsymbol\lambda$. Hence, the minimax theorem \cite{VonNeumann1971,DPa95} can be utilised to change the optimisation order as
\begin{eqnarray}
\nonumber \max_{\boldsymbol\lambda}&\Big\{-\delta\boldsymbol\lambda^\top\boldsymbol\lambda+2\boldsymbol\lambda^\top\mathbf{1}+\min_{\boldsymbol\mu}-\boldsymbol\mu^\top\mathbf{u}\Big\}\\
&\text{s.t. } \boldsymbol\mu\geq 0, \norm{\boldsymbol\mu\boldsymbol\mu^\top}_{p,q}\leq 1
\label{fobj}
\end{eqnarray}
The resulting optimisation problem above has thus become compatible with that of Eq. \ref{pi1} of the non-localised MKL approach with the difference that in the proposed localised method the dimensionality of the kernel weight vector, i.e. $\boldsymbol\mu$, is increased by a factor of number of clusters, i.e. $C$. Since the structure of the learning problem corresponding to the proposed localised approach in Eq. \ref{fobj} matches that of Eq. \ref{pi1}, following a similar optimisation approach as that advocated in \cite{9535490}, $\boldsymbol\mu$ is derived as (a complete derivation of $\boldsymbol\mu$ is provided in the Appendix \ref{derv})
\begin{eqnarray}
\nonumber \boldsymbol\mu = \mathbf{\bar{u}}/\sqrt{\norm{\bar{\mathbf{u}}}_p\norm{\bar{\mathbf{u}}}_q}
\end{eqnarray}
\noindent where 
\begin{eqnarray}
\bar{\mathbf{u}}=\mathbf{u}\odot\Big(\boldsymbol\mu^{p-2}/\norm{\boldsymbol\mu}_p^p+\boldsymbol\mu^{q-2}/\norm{\boldsymbol\mu}_q^q\Big)^{-1}
\label{optsol}
\end{eqnarray}
\noindent where $\odot$ denotes a Hadamard (element-wise) multiplication. Once $\boldsymbol\mu$ is calculated, the next step is to maximise the objective function w.r.t. $\boldsymbol\lambda$. This may be readily realised by following the standard approach to solve for the regularised Fisher null projection \cite{9055448} with an effective kernel matrix of $\mathbf{K}=\sum_{c=1}^C\sum_{g=1}^G\mu_{cg}\mathbf{K}_{cg}$. Setting the partial derivative of the cost function w.r.t. $\boldsymbol\lambda$ to zero yields
\begin{eqnarray}
\boldsymbol\lambda = \Big(\delta \mathbf{I}+\sum_{c=1}^C\sum_{g=1}^G\mu_{cg}\mathbf{K}_{cg}\Big)^{-1}\mathbf{1}
\label{omegasingle2}
\end{eqnarray}
\begin{algorithm}[t]
\footnotesize
\caption{Localised Joint $\ell_{p,q}$-norm Multiple Kernel Null-Space Fisher Method}
\label{localMKL}
\begin{algorithmic}[1]
\State \textit{Estimate} $p_c(\mathbf{x}_i)$'s; $i=1,\dots,n$;  $c=1,\dots,C$.
\State \textit{Construct kernel matrices}: $\mathbf{K}_{cg}$'s; $c=1,\dots,C$;  $g=1,\dots,G$.
\State \textit{Initialisation:} $\boldsymbol\lambda = \big(\delta \mathbf{I}+\sum_{c=1}^C\sum_{g=1}^G (CG)^{\frac{-p-q}{2pq}}\mathbf{K}_{cg}\big)^{-1}\mathbf{1}$
\Repeat
\For {$c=1,\dots,C$}
\State $\mathbf{u}_c=[\boldsymbol\lambda^\top\mathbf{K}_{c1}\boldsymbol\lambda,\boldsymbol\lambda^\top\mathbf{K}_{c2}\boldsymbol\lambda,\dots,\boldsymbol\lambda^\top\mathbf{K}_{cG}\boldsymbol\lambda]^\top$
\EndFor
\State $\mathbf{u}=[\mathbf{u}_1^\top,\mathbf{u}_2^\top,\dots,\mathbf{u}_C^\top]^\top$
\State $\bar{\mathbf{u}}=\mathbf{u}\odot\Big(\boldsymbol\mu^{p-2}/\norm{\boldsymbol\mu}_p^p+\boldsymbol\mu^{q-2}/\norm{\boldsymbol\mu}_q^q\Big)^{-1}$
\State $\boldsymbol\mu = \mathbf{\bar{u}}/\sqrt{\norm{\bar{\mathbf{u}}}_p\norm{\bar{\mathbf{u}}}_q}$

\State $\boldsymbol\lambda = \Big(\delta \mathbf{I}+\sum_{c=1}^C\sum_{g=1}^G\mu_{cg}\mathbf{K}_{cg}\Big)^{-1}\mathbf{1}$
\Until{convergence}
\State Output: $\boldsymbol{\lambda}$ and $\boldsymbol\mu$
\end{algorithmic}
\normalsize
\end{algorithm}
Note that the relation above gives $\boldsymbol\lambda$ in terms of $\boldsymbol\mu$ which is itself a function of $\boldsymbol\lambda$. Denoting $\Big(\delta \mathbf{I}+\sum_{c=1}^C\sum_{g=1}^G\mu_{cg}\mathbf{K}_{cg}\Big)^{-1}\mathbf{1}\coloneqq h(\boldsymbol\lambda)$, the optimal $\boldsymbol\lambda$ should then fulfil $\boldsymbol\lambda=h(\boldsymbol\lambda)$. In order to determine $\boldsymbol\lambda$ satisfying $\boldsymbol\lambda=h(\boldsymbol\lambda)$, a fixed-point iteration method \cite{NM} is applied. Algorithm \ref{localMKL} summarises the approach described above where $\boldsymbol\lambda$ is initialised using uniform kernel weights with a unit $\ell_{p,q}$-norm. Once $\boldsymbol\lambda$ and $\boldsymbol\mu$ are computed, an observation $\mathbf{y}$ can be mapped onto the localised Fisher null subspace as
\begin{eqnarray}
\nonumber f(\mathbf{y})=\sum_{c=1}^Cp_c(\mathbf{y})\sum_{g=1}^G\mu_{cg}\Big[\kappa_g(\mathbf{y},\mathbf{x}_1)p_c(\mathbf{x}_1),\\
\dots,\kappa_g(\mathbf{y},\mathbf{x}_n)p_c(\mathbf{x}_n)\Big]\boldsymbol\lambda
\end{eqnarray}

\noindent The following proposition characterises the convergence properties of the proposed localised matrix-norm multiple kernel learning approach summarised in Algorithm \ref{localMKL}.
\begin{prop}
For an adequately large $\delta$, the proposed approach will converge to a single point independent of a particular initilisation for $\boldsymbol\lambda$.
\label{prop1}
\end{prop}
\noindent Consult Appendix \ref{prop1proof} for a proof.

\section{Theoretical Analysis}
\label{TA}
In this section, a theoretical analysis of the proposed localised $\ell_{p,q}$-norm one-class MKL approach based on Rademacher complexities is conducted which facilitates deriving a generalisation error bound for the proposed method. Furthermore, the derived Rademacher complexity bound enables a theoretical comparison between the proposed approach and some other alternative methods. 

The hypothesis set for the proposed localised $\ell_{p,q}$-norm one-class MKL method with $C$ clusters and $G$ kernels may be expressed as
\begin{eqnarray}
\label{DH}
\nonumber &&H_{CG}^{p,q}=\Big\{f:\mathcal{X}\rightarrow\mathbb{R}\Big|f(\mathbf{y})=\sum_cp_c(\mathbf{y})\sum_g\mu_{cg}\big[\kappa_g(\mathbf{y},\mathbf{x}_1)\\\nonumber &&p_c(\mathbf{x}_1),\dots,\kappa_g(\mathbf{y},\mathbf{x}_n)p_c(\mathbf{x}_n)\big]\boldsymbol\lambda,\\ \nonumber &&\boldsymbol\lambda^\top\big(\sum_c\sum_g\mu_{cg}
\mathbf{K}_{cg}\big)\boldsymbol\lambda\leq \Lambda^2,\norm{\boldsymbol\mu\boldsymbol\mu^\top}_{p,q}\leq 1\Big\}\\
\end{eqnarray}
where the constraint $\boldsymbol\lambda^\top\big(\sum_c\sum_g\mu_{cg}
\mathbf{K}_{cg}\big)\boldsymbol\lambda\leq \Lambda^2$ follows from the equivalency between a Tikhonov and an Ivanov regularisation for some $\Lambda$ \cite{JMLR:v12:kloft11a}. The empirical Rademacher complexity \cite{MohriRostamizadehTalwalkar18} of the hypothesis class $H_{CG}^{p,q}$ for samples $\mathbf{x}_i$'s is
\begin{eqnarray}
\mathcal{R}(H_{CG}^{p,q})=\frac{1}{n}\mathbb{E}_{\boldsymbol\sigma}\Big[\mathrm{sup}_{f\in H_{CG}^{p,q}}\sum_{i} \sigma_if(\mathbf{x}_i)\Big]
\end{eqnarray}
where $\sigma_i$'s are uniform independent random variables with values in $\{-1,+1\}$ called Rademacher variables and $\boldsymbol\sigma$ is a vector collection of $\sigma_i$'s. The Rademacher complexity measures the richness of the function class $H_{CG}^{p,q}$ by quantifying how well the class of functions under analysis is capable of correlating with random labels.
\begin{prop}
The empirical Rademacher complexity of the proposed approach is bounded as
 \begin{eqnarray}
 \mathcal{R}(H_{CG}^{p,q})\leq \frac{\Lambda r}{n}(CG)^{\frac{1}{2}-\frac{1}{4p}-\frac{1}{4q}}\Bigg(\sum_c\sum_ip_c^2(\mathbf{x}_i)\Bigg)^{1/2}
 \label{deriverade}
 \end{eqnarray}
  \label{prop2}
\end{prop}
\noindent A proof is provided in Appendix \ref{prop2proof}.

Using the Rademacher complexity, one may characterise the generalisation performance of the proposed algorithm as follows.

\begin{prop}[Generalisation error bound]
Assuming that the kernel function is bounded by $r^2$ and the squared error loss is bounded by $B_l$ and is $\mu$-Lipschitz for some $\mu$, then the following inequality holds with confidence larger than $1-\Delta$ over samples of size $n$ for all classifiers $f\in\mathcal{H}_{CG}^{p,q}$
\end{prop}
\begin{eqnarray}
\nonumber &&\varepsilon(f)\leq \frac{1}{n}\sum_i (f(\mathbf{x}_i)-1)^2+\\
\nonumber &&\frac{\Lambda r}{n}(CG)^{\frac{1}{2}-\frac{1}{4p}-\frac{1}{4q}}\Bigg(\sum_c\sum_ip_c^2(\mathbf{x}_i)\Bigg)^{1/2}+3B_l\sqrt{\frac{\ln(2/\Delta)}{2n}}\\
\label{geb}
\end{eqnarray}
where $\varepsilon(f)\coloneqq\mathbb{E}\Big[(\hat{f}(\mathbf{y})-f(\mathbf{y}))^2\Big]$, and $\hat{f}(.)$ denotes the true label of an observation.

\textbf{Proof} The proof follows by plugging in the Rademacher complexity bound and the empirical loss function of the proposed approach into Theorem 11.3 in \cite{MohriRostamizadehTalwalkar18}.$\square$

The generalisation error bound in Eq. \ref{geb} suggests a trade-off between reducing the empirical error over the training set and controlling the Rademacher complexity of $\mathcal{H}_{CG}^{p,q}$. As suggested by Eq. \ref{deriverade}, the empirical Rademacher complexity for the hypothesis set $\mathcal{H}_{CG}^{p,q}$ may be controlled using $\Lambda$ which depends on $\delta$ (see Appendix \ref{depil} for a proof). As a result, in order to optimise the generalisation error bound, one needs to set $\theta$ (or equivalently $\delta$) so as to maintain a trade-off between different terms in the upper bound given in Eq. \ref{geb}. In practice, a validation set independent from the training set may be used to tune $\theta$ so that a performance metric is optimised.

\subsection{Other regularisation constraints}
In this study, we considered a \textit{joint} $\ell_{p,q}$-norm constraint on the collection of kernel weights from all the existing clusters. Other closely related alternatives for norm regularisation include: (1) a joint vector $\ell_p$-norm; (2) a \textit{disjoint} vector $\ell_p$-norm; or (3) a \textit{disjoint} matrix $\ell_{p,q}$-norm. For the joint vector $\ell_p$-norm regularisation, the constraint is expressed as $\norm{\boldsymbol\mu}_p\leq1$ while for the disjoint vector-norm regularisation, the constraint set is defined as $\norm{\boldsymbol\mu_c}_p\leq1$, for $c=1,\dots, C$. Regarding the disjoint $\ell_{p,q}$-norm regularisation, the constraints are $\norm{\boldsymbol\mu_c\boldsymbol\mu_c^\top}_{p,q}\leq1$, for $c=1,\dots, C$. As formally stated in the following proposition, the empirical Rademacher complexity bound of the proposed joint matrix-norm approach is lower than those of the joint vector-norm, disjoint vector-norm, and the disjoint matrix-norm class.
 \begin{prop}
The empirical Rademacher complexity of the proposed localised joint $\ell_{p,q}$-norm ($1\leq q\leq p$) one-class MKL approach is
\begin{enumerate}[label=(\roman*)]
\item lower than the localised joint vector-norm function class by a factor of $(CG)^{\frac{1}{4p}-\frac{1}{4q}}$
\item lower than the localised disjoint vector-norm function class by a factor of $C^{-\frac{1}{4p}-\frac{1}{4q}}G^{\frac{1}{4p}-\frac{1}{4q}}$,
\item lower than the localised disjoint matrix-norm function class by a factor of $C^{-\frac{1}{4p}-\frac{1}{4q}}$
\end{enumerate}
\label{prop3}
\end{prop}
\noindent A proof is provided in Appendix \ref{prop3proof}.

\subsection{Runtime Complexity}
During training, we incur $\mathcal{O}(CGn^2)$ for computing kernel matrices, and during each training iteration $\mathcal{O}(Cn^2)$ for computing vector $\mathbf{u}$ while for solving the linear equation $\Big(\delta \mathbf{I}+\sum_{c=1}^C\sum_{g=1}^G\mu_{cg}\mathbf{K}_{cg}\Big)\boldsymbol\lambda = \mathbf{1}$ via incremental Cholesky factorisation using Sherman’s March algorithm \cite{0898714141} and forward-back substitution we need $\mathcal{O}(n^2)$. The complexity at the test phase is $\mathcal{O}(CGn)$.

\section{Experiments and Results}
\label{ER}
In this section, the results of evaluating the proposed approach on different datasets are presented and compared against existing methods. The rest of this section is structured as detailed below.
\begin{itemize}
    \item Section \ref{ID}, provides the details of our implementation.
    \item In Section \ref{vis}, we visualise the local kernel weights inferred for some sample classes to confirm the possibility of inferring locally adaptive kernel weights using the proposed method.
    \item In Section \ref{GO}, the results of evaluating the proposed approach on one-class general object classification for abnormality and novelty detection on the Caltech256 \cite{256} and the 1001 abnormal objects \cite{6618951} datasets are presented and compared against other methods.
    \item Section \ref{OCFP} presents the results of the unseen face PAD on four widely used datasets in a zero-shot evaluation scenario and compares the proposed approach with the state-of-the-art techniques.
\end{itemize}

\begin{figure}[t]
\centering
\includegraphics[scale=.29]{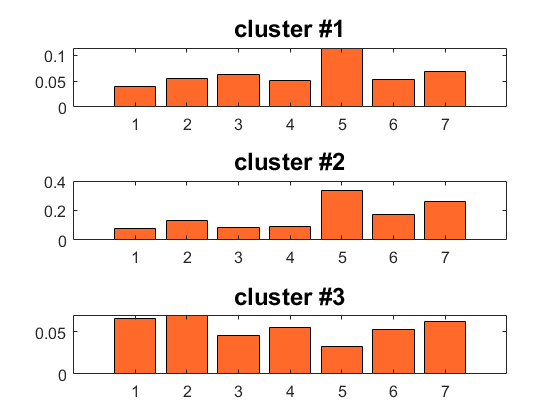}
\includegraphics[scale=.29]{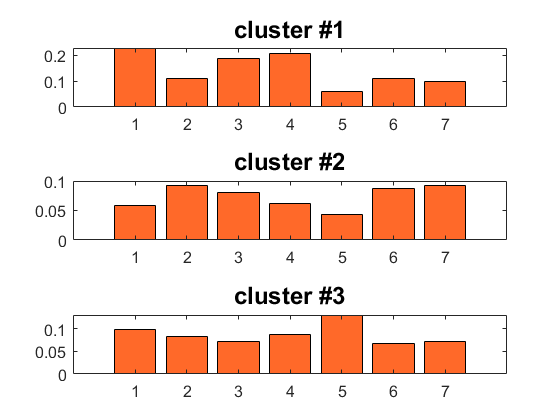}
\caption{Sample local weights for seven kernels in 3 different clusters for two sample classes (left and right columns) from the Caltech256 dataset.}
\label{localweights}
\end{figure}

\subsection{Details of implementation}
\label{ID}
In the following experiments, we use an RBF kernel function and set its width to half of the mean pairwise Euclidean distance over training observations. $\delta$ is selected from $\{10^{-3}, 10^{-2}, 10^{-1}, 1, 10, 10^2\}\times n$ while $p$ and $q$ are selected from $\{32/31, 16/15, 8/7, 4/3, 2, 4, 8, 10\}$. On all datasets, the parameters above are set using the data from classes other than the target class under consideration. The number of genuine data clusters $C$ is set to 3 in all the following experiments. We form the clusters using the kernel k-means algorithm and derive $p_c(\mathbf{x}_i)$'s using a softmax function operating on the distances from cluster centres.
\subsection{Effect of locality}
\label{vis}
The objective of the current study is to infer locally adaptive kernel weights. In order to visualise and confirm that different weights can be inferred for different data clusters, we randomly pick two classes from the Caltech256 dataset \cite{256}, and following \cite{9535490}, construct seven kernels using deep representations derived from the pre-trained convolutional neural networks of Resnet50 \cite{He2016DeepRL}, Googlenet \cite{7298594}, Alexnet \cite{NIPS2012_c399862d}, Vgg16 \cite{Simonyan14c}, Densenet201 \cite{8099726}, Mobilenetv2 \cite{8578572}, and Nasnetlarge \cite{8579005}. We set the number of clusters to three and run the proposed algorithm to infer local kernel weights. The weights inferred for the seven kernel matrices used are visualised in Fig. \ref{localweights}. As it can be seen from the figure, for both object classes, a specific kernel can be weighted differently in different clusters. As such, the proposed approach is effective is inferring locally adaptive kernel weights. The utility of the inferred local weights for one-class classification is analysed in the following sections.
\subsection{One-class general object classification}
\label{GO}
In the first set of experiments, the proposed approach is evaluated for one-class classification of general object images. For this purpose, two experiments for novelty and abnormality detection are conducted on the Caltech256 and 1001 abnormal objects datasets, discussed next.
\subsubsection{Novelty detection}
The Caltech 256 dataset \cite{256} is commonly used to test novelty detection algorithms. The dataset incorporates images of objects from 256 different classes giving rise to 30607 images in total. For novelty detection, different protocols have been introduced to convert the multi-class dataset into one suitable for one-class novelty detection. So as to conduct a fair comparison with other approaches, in this work, we follow one of the widely used evaluation protocols introduced in \cite{DBLP:journals/corr/abs-1801-05365} where a single class is presumed as the target/normal class and the learned model is then evaluated by considering objects from all the remaining 255 classes as novel objects. The experiment is repeated for the first 40 classes of the database.
\subsubsection{Abnormality detection}
For abnormality detection, one of the frequently employed image datasets is the 1001 Abnormal Objects database \cite{6618951}. The dataset includes 1001 image samples of six different object classes from the PASCAL database \cite{everingham2009the}. The goal on this database is to classify each image as either normal or abnormal. While the abnormal samples correspond to 6 different categories and are annotated using Amazon Mechanical Turk, the normal instances are obtained from the PASCAL dataset from each corresponding object class.

On the Caltech256 and 1001 Abnormal object databases, we use the generally known AUC performance metric to compare the performance of the proposed method to those of existing methods. The AUC provides a performance statistic independent of a particular threshold by measuring the area under the Receiver Operating Characteristic curve (ROC), which contrasts the true positive rate against the false positive rate at varying thresholds. In order to specifically gauge the performance benefits brought by the proposed \textit{localised} MKL approach compared to the non-localised matrix-regularised MKL \cite{9535490}, we use the features obtained from the seven deep pre-trained CNN's of Resnet50 \cite{He2016DeepRL}, Googlenet \cite{7298594}, Alexnet \cite{NIPS2012_c399862d}, Vgg16 \cite{Simonyan14c}, Densenet201 \cite{8099726}, Mobilenetv2 \cite{8578572}, and Nasnetlarge \cite{8579005} as suggested in \cite{9535490} to construct $21$ localised kernel matrices ($7$ representations $\times$ $3$ clusters) and compare the performance of the proposed method to its non-localised variant as well as other methods from the literature. The results of this set of experiments are tabulated in Table \ref{ImageNovelty} and Table \ref{ImageAbnormality} for the Caltech256 and 1001 Abnormal Objects datasets, respectively. From the tables, the following observations may be made. On the Caltech256 dataset, while the non-localised MKL approach achieves an average AUC of $99.6\%$, the proposed localisation improves this performance and yields a perfect classification for novelty detection, ranking the best for novelty detection on the Caltech256 database. On the 1001 Abnormal Objects database, while the average AUC of the non-localised method is $99.2\%$, the proposed localised MKL approach obtains an average AUC of $99.4\%$, performing better than other alternatives.

\begin{table}
\renewcommand{\arraystretch}{1.2}
\caption{Performances of different methods on the Caltech256 database for novelty detection in terms of AUC (MEAN$\pm$STD$\%$).}
\label{ImageNovelty}
\centering
\begin{tabular}{lc}
\hline
\textbf{Approach} & \textbf{AUC}\\
\hline
OCNN-AlexNet \cite{DBLP:journals/corr/abs-1802-06360} &$82.6\pm15.3$\\
OCNN-VGG16 \cite{DBLP:journals/corr/abs-1802-06360} &$88.5\pm14.4$\\
Autoencoder \cite{1640964}& $62.3\pm7.2$\\
OCSVM-VGG16 \cite{8721681} &$90.2\pm5.0$\\
DOC-VGG16 \cite{8721681} &$98.1\pm2.2$\\
DOC-AlexNet \cite{8721681} &$95.9\pm2.1$\\
Multiple Kernel-SVDD \cite{9535490}&$99.1\pm0.7$\\
Multiple Kernel-SVDD-Slim \cite{9535490}&$99.2\pm0.6$\\
Multiple Kernel-OCSVM \cite{9535490}&$99.1\pm0.8$\\
Multiple Kernel-OCSVM-Slim \cite{9535490}&$99.1\pm0.8$\\
$\ell_{p,q}$-MKL \cite{9535490} &$99.6\pm0.5$\\
Localised MKL (this work) &$\mathbf{100\pm0}$\\
\hline
\end{tabular}
\end{table}

\begin{table}
\renewcommand{\arraystretch}{1.2}
\caption{Performances of different methods on the Abnormality-1001 database for abnormality detection in terms of AUC (MEAN$\pm$STD$\%$).}
\label{ImageAbnormality}
\centering
\begin{tabular}{lc}
\hline
\textbf{Approach} & \textbf{AUC}\\
\hline
Graphical Model \cite{6618951} & $87.0\pm n.a.$ \\
Adj. Graphical Model \cite{6618951}& $91.1\pm n.a.$\\
OCNN \cite{DBLP:journals/corr/abs-1802-06360}& $88.5\pm1.4$\\
Autoencoder \cite{1640964} & $67.4\pm1.2$\\
OC-CNN \cite{8586962}& $84.3\pm n.a.$\\
DOC \cite{8721681} & $95.6\pm3.1$ \\
Multiple Kernel-SVDD \cite{9535490}&$92.3\pm0.2$\\
Multiple Kernel-SVDD-Slim \cite{9535490}&$94.1\pm0.1$\\
Multiple Kernel-OCSVM \cite{9535490}&$91.4\pm0.3$\\
Multiple Kernel-OCSVM-Slim \cite{9535490}&$93.9\pm0.1$\\
$\ell_{p,q}$-MKL \cite{9535490} &$99.2\pm0.2$\\
Localised MKL (this work) &$\mathbf{99.4\pm0.4}$\\
\hline
\end{tabular}
\end{table}

\subsection{One-class unseen face PAD}
\label{OCFP}
In this set of experiments, the proposed approach is evaluated for the unknown face PAD and compared to other methods that follow a similar zero-shot unseen PA detection setting. For this purpose, we use four commonly used face PAD databases briefly introduced below.
\subsubsection{The Replay-Mobile database \cite{Costa-Pazo_BIOSIG2016_2016}}
incorporates 1190 video sequences of genuine and attack observations from 40 subjects that are captured by two different video cameras under various lighting conditions. In this dataset, three disjoint subsets exist which are to be used for training, development, and testing purposes as well as an additional enrolment subset.
\subsubsection{The OULU-NPU database \cite{7961798}}
includes 4950 attack and genuine videos corresponding to 55 individuals recorded with 6 different cameras in 3 sessions subject to various lighting conditions and background settings. The dataset evaluation protocol requires PA detection subject to previously unseen video acquisition settings, unseen attack types and also unseen camera sensors. The videos are separated into three subject-disjoint groups as train, development, and test sets. For testing, 4 evaluation protocols are considered among which Protocol IV is the most difficult one and is followed in this work. \subsubsection{The MSU-MFSD dataset \cite{7031384}}
incorporates 440 videos recorded from either video or photo attack trials corresponding to 55 individuals which are captured using 2 different devices. The subset of the dataset which is publicly available, includes videos corresponding to 35 subjects. The dataset is partitioned to 2 sets for training and testing purposes with no overlap between the subjects from the two sets.
\subsubsection{The Replay-Attack dataset \cite{6313548}}
incorporates 1300 videos of genuine and attack data from fifty subjects. Attack data are generated by means of a mobile phone, an HD iPad screen, or using a printed image. The database is randomly divided to three subject-disjoint partitions for training, development, and testing purposes.

The ISO metrics \cite{BSISO-IEC30107-3-2017} used to gauge presentation attack detection system performance are: 1) BPCER: bona fide presentation classification error rate that measures the ratio of genuine samples which are incorrectly classified; and 2) APCER: attack presentation classification error rate that reflects the ratio of incorrectly classified PA's realised with a similar PA instrument species. Based on the maximum APCER over all attack instruments, the presentation attack detection system performance may be summarised in terms of ACER, i.e. the Average Classification Error Rate as
\begin{eqnarray}
ACER = \Big(BPCER+\max_{PAIS}APCER_{PAIS}\Big)/2
\end{eqnarray}
In addition to the aforementioned metrics and to enable a comparison of the performance of the proposed method to other approaches, we also report the performance of the proposed method in terms of the Area Under the ROC Curve (AUC) and Half Total Error Rate (HTER). The kernels used in this set of experiments are those suggested in \cite{9535490} which have been observed to be effective for face presentation attack detection. These kernels correspond to deep CNN features extracted using Resnet50 \cite{He2016DeepRL}, Googlenet \cite{7298594}, and Vgg16 \cite{Simonyan14c} pre-trained models from four different facial regions, giving rise to $36$ kernels in the proposed localised approach: $3\mathrm{(CNN's)}\times4\mathrm{(regions)}\times3\mathrm{(clusters)}$. Using similar deep representations as those suggested in \cite{9535490} enables a direct comparison between the proposed localised MKL method and its non-localised variant. The results of this set of experiments are provided in Table \ref{RM}, Table \ref{ONC}, Table \ref{MSU}, and Table \ref{RA}, for the Replay-Mobile \cite{Costa-Pazo_BIOSIG2016_2016}, Oulu-NPU \cite{7961798}, MSU-MFSD \cite{7031384}, and Replay-Attack \cite{6313548} databases, respectively. The following observations from the tables are in order. On the Replay-Mobile database, in an unknown face PAD setting, the non-localised MKL method of \cite{9535490} obtains a HTER of $6.7\%$. The proposed localised MKL approach reduces the error rate to $5.6\%$ accounting to more than $15\%$ improvement in terms of HTER. The proposed localised approach also outperforms other multiple kernel learning systems operating on SVM formulations. The proposed approach also yields a better performance compared to the weighted classifier fusion method presented in \cite{8987326} and also better than the method of \cite{FATEMIFAR2021107696}. On the Oulu-NPU protocol IV, a similar observation may be made. That is, the proposed localised learning approach improves upon its non-localised variant. In particular, while the average APCER of the proposed approach is $3.3\%$, for the non-localised version the corresponding average APCER is $5.0$ which amounts to more than $30\%$ improvement in terms of APCER. The overall performance of the proposed method on the Oulu-NPU dataset measured as ACER is better than other methods evaluated in an unknown face spoofing detection scenario. For the MSU-MFSD and the Replay-Attack databases, the proposed approach provides a perfect separation between bona fide and attack samples as may be observed from the corresponding tables.

\begin{table}[t]
\renewcommand{\arraystretch}{1.2}
\caption{Performances of different methods in terms of Half Total Error Rate ($\%$) for the unknown face PAD on the Replay-Mobile database.}
\label{RM}
\centering
\begin{tabular}{lcc}
\hline
\textbf{Approach} & \textbf{HTER}\\
\hline
The work in \cite{FATEMIFAR2021107696} & $8.5$\\
Weighted-Fusion \cite{8987326} & $9.9$\\
GoogleNet+MD \cite{8682253} & $13.7$\\
The work in \cite{9303455} &$11.8$\\
Multiple Kernel-SVDD \cite{9535490}&$11.1$\\
Multiple Kernel-SVDD-Slim \cite{9535490}&$7.6$\\
Multiple Kernel-OCSVM \cite{9535490}&$10.8$\\
Multiple Kernel-OCSVM-Slim \cite{9535490}&$8.1$\\
$\ell_{p,q}$-MKL \cite{9535490} &$6.7$\\
Localised MKL (this work) &$\mathbf{5.6}$\\
\hline
\end{tabular}
\end{table}

\begin{table}[t]
\renewcommand{\arraystretch}{1.2}
\caption{Performances of different methods in terms of BPCER, APCER, and ACER ($\%$) for the unknown face PAD on the Oulu-NPU database protocol IV (mean$\pm$std $\%$).}
\label{ONC}
\centering
\begin{tabular}{lcccc}
\hline
\textbf{Approach} &\textbf{BPCER}&\textbf{APCER}& \textbf{ACER}\\
\hline
SAPLC \cite{9056824} &$6.6\pm5.5$&$11.9\pm6.9$&$9.3\pm4.4$\\
OCA-FAS \cite{QIN2020384} &$5.9\pm4.5$&$2.3\pm2.5$&$4.1\pm2.7$\\
The work in \cite{feng2020learning2} &$1.7\pm2.6$&$5.8\pm4.9$& $3.7\pm2.1$\\
The work in \cite{8737949} &$9.7\pm4.8$&$11.3\pm3.9$&$9.8\pm4.2$\\
The work in \cite{9303455} &$0.8\pm2.0$&$11.6\pm13.6$&$6.2\pm6.8$\\
Multiple Kernel-SVDD \cite{9535490}&$2.5\pm4.2$&$11.7\pm10.8$&$7.1\pm6.2$\\
Multiple Kernel-SVDD-Slim \cite{9535490}&$1.7\pm4.1$&$10.8\pm8.0$&$6.2\pm4.4$\\
Multiple Kernel-OCSVM \cite{9535490}&$2.5\pm4.2$&$13.3\pm11.7$&$7.9\pm6.4$\\
Multiple Kernel-OCSVM-Slim \cite{9535490}&$1.7\pm4.1$&$10.8\pm8.0$&$6.2\pm4.4$\\
$\ell_{p,q}$-MKL \cite{9535490}&$0\pm0$&$5.0\pm4.5$&$2.5\pm2.2$\\
Localised MKL (this work) &$0\pm0$&$3.3\pm4.1$&$\mathbf{1.6\pm2.0}$\\
\hline
\end{tabular}
\end{table}

\begin{table}[t]
\renewcommand{\arraystretch}{1.2}
\caption{Performances of different methods in terms of AUC ($\%$) for the unknown face PAD on on the MSU-MFSD database.}
\label{MSU}
\centering
\begin{tabular}{lcc}
\hline
\textbf{Method} & \textbf{AUC}\\
\hline
IMQ+OCSVM \cite{7984788} &$67.7$\\
BSIF+OCSVM \cite{7984788} & $75.6$\\
LBP+NN \cite{8698574} &$81.6$\\
LBP+GMM \cite{8698574} &$81.3$\\
LBP+OCSVM \cite{8698574} &$84.5$\\
LBP+AE \cite{8698574} &$87.6$\\
DTL \cite{8953776} &$93.0$\\
The work in \cite{9303455} & $\mathbf{100}$\\
Multiple Kernel-SVDD \cite{9535490}&$\mathbf{100}$\\
Multiple Kernel-SVDD-Slim \cite{9535490}&$\mathbf{100}$\\
Multiple Kernel-OCSVM \cite{9535490}&$\mathbf{100}$\\
Multiple Kernel-OCSVM-Slim \cite{9535490}&$\mathbf{100}$\\
$\ell_{p,q}$-MKL \cite{9535490}&$\mathbf{100}$\\
Localised MKL (this work) &$\mathbf{100}$\\
\hline
\end{tabular}
\end{table}

\begin{table}[t]
\renewcommand{\arraystretch}{1.2}
\caption{Performances of different methods in terms AUC ($\%$) for the unknown face PAD on on the Replay-Attack database.}
\label{RA}
\centering
\begin{tabular}{lcc}
\hline
\textbf{Method} & \textbf{AUC}\\
\hline
IMQ+OCSVM \cite{7984788} &$80.7$\\
BSIF+OCSVM \cite{7984788} & $81.9$\\
LBP+NN \cite{8698574} &$91.2$\\
LBP+GMM \cite{8698574} &$90.1$\\
LBP+OCSVM \cite{8698574} &$87.9$\\
LBP+AE \cite{8698574} &$86.1$\\
DTL \cite{8953776} &$99.8$\\
MD \cite{8682253} &$99.7$\\
The work in \cite{9303455} & $\mathbf{100}$\\
Multiple Kernel-SVDD \cite{9535490}&$\mathbf{100}$\\
Multiple Kernel-SVDD-Slim \cite{9535490}&$\mathbf{100}$\\
Multiple Kernel-OCSVM \cite{9535490}&$\mathbf{100}$\\
Multiple Kernel-OCSVM-Slim \cite{9535490}&$\mathbf{100}$\\
$\ell_{p,q}$-MKL \cite{9535490}&$\mathbf{100}$\\
Localised MKL (this work) &$\mathbf{100}$\\
\hline
\end{tabular}
\end{table}

\section{Conclusion}
\label{conc}
We considered the unseen face PAD problem in the context of one-class multiple kernel learning. While earlier MKL studies failed to explicitly take into the variability in bona fide samples in learning kernel weights, we presented a localised one-class MKL algorithm which provided locally adaptive kernel weights. We formulated the corresponding problem as a convex optimisation task and presented an effective method for its optimisation. The results of evaluating the proposed method on general object image and face spoofing data verified its efficacy in detection presentation attacks by solely training on genuine samples while a theoretical analysis of the proposed localised multiple kernel learning approach based on Rademacher complexities illustrated its benefits compared to some other alternatives. 

\section*{Acknowledgements}
This research is supported by The Scientific and Technological Research Council of Turkey (TÜBİTAK) under the grant no 121E465.
\bibliographystyle{IEEEtran}
\bibliography{IEEEexample.bib}

\appendices
\section{Derivation of $\boldsymbol\mu$}
\label{derv}
As noted in Section IV.D, the theorem of minimax can be utilised to change the minimisation and maximisation order in Eq. (16) to obtain
\begin{eqnarray}
\max_{\boldsymbol\lambda} -\delta\boldsymbol\lambda^\top\boldsymbol\lambda+2\boldsymbol\lambda^\top\mathbf{1}+\min_{\boldsymbol\mu\in \mathcal{J}}-\boldsymbol\mu^\top\mathbf{u}
\label{switched}
\end{eqnarray}
\noindent where $\mathcal{J}=\{\boldsymbol\mu\Big| \boldsymbol\mu \geq 0, \norm{\boldsymbol\mu\boldsymbol\mu^\top}_{p,q} \leq 1 \}$. 
Let us form the Lagrangian of the minimisation sub-problem in Eq. \ref{switched} as
\begin{eqnarray}
\mathcal{L} = \gamma(\norm{\boldsymbol\mu\boldsymbol\mu^\top}_{p,q}-1)-(\mathbf{u}+\boldsymbol{\chi})^\top\boldsymbol\mu
\end{eqnarray}
where $\boldsymbol\chi\geq0$ and $\gamma\geq0$ are non-negative Lagrange multipliers. The Karush–Kuhn–Tucker (KKT) optimality criteria are:
\begin{subequations}
\begin{align}
 &\nabla_{\boldsymbol\mu} \mathcal{L} = 0 \label{eq:subeq1}\\
 &\boldsymbol\chi^\top\boldsymbol\mu = 0 \label{eq:subeq2}\\
 &\boldsymbol\mu\geq 0 \label{eq:subeq3}\\
 &\gamma(\norm{\boldsymbol\mu\boldsymbol\mu^\top}_{p,q}-1)=0\label{eq:subeq4}
\end{align}
\end{subequations}
Using \eqref{eq:subeq1} and \eqref{eq:subeq3} one obtains 
\begin{eqnarray}
-\mathbf{u}-\boldsymbol\chi+\gamma\Big[\frac{\norm{\boldsymbol\mu}_p\norm{\boldsymbol\mu}_q}{\norm{\boldsymbol\mu}_p^p}\boldsymbol\mu^{p-1}+\frac{\norm{\boldsymbol\mu}_q\norm{\boldsymbol\mu}_p}{\norm{\boldsymbol\mu}_q^q}\boldsymbol\mu^{q-1} \Big]=0
\label{pos}
\end{eqnarray}
By inspecting the minimisation subproblem in Eq. \ref{switched}, one understands that the elements of $\boldsymbol\mu$ should be maximised for optimality. As the matrix norm is a convex function of $\boldsymbol\mu$, maximisation of the components of $\boldsymbol\mu$ maximises $\norm{\boldsymbol\mu\boldsymbol\mu^\top}_{p,q}$ whose maximum occurs on the border of the set $\mathcal{J}$ which is given as $\norm{\boldsymbol\mu\boldsymbol\mu^\top}_{p,q}=1$. As we have $\norm{\boldsymbol\mu\boldsymbol\mu^\top}_{p,q}=\norm{\boldsymbol\mu}_p\norm{\boldsymbol\mu}_q$ \cite{8259375}, at the optimal point it holds $\norm{\boldsymbol\mu}_p\norm{\boldsymbol\mu}_q=1$. Eq. \ref{pos} can now be expressed as
\begin{eqnarray}
\mathbf{u}+\boldsymbol\chi=\gamma\Big(\frac{\boldsymbol\mu^{p-1}}{\norm{\boldsymbol\mu}_p^p}+\frac{\boldsymbol\mu^{q-1}}{\norm{\boldsymbol\mu}_q^q}\Big)=\gamma\boldsymbol\mu\odot\Big(\frac{\boldsymbol\mu^{p-2}}{\norm{\boldsymbol\mu}_p^p}+\frac{\boldsymbol\mu^{q-2}}{\norm{\boldsymbol\mu}_q^q}\Big)
\label{interm1}
\end{eqnarray}
Eq. \ref{interm1} may be rewritten as
\begin{eqnarray}
\bar{\mathbf{u}}+\bar{\boldsymbol\chi}=\gamma\boldsymbol\mu
\label{sumeq}
\end{eqnarray}
where $\bar{\mathbf{u}}$ and $\bar{\boldsymbol\chi}$ are defined as
\begin{eqnarray}
\bar{\mathbf{u}}=\mathbf{u}\odot\Big(\frac{\boldsymbol\mu^{p-2}}{\norm{\boldsymbol\mu}_p^p}+\frac{\boldsymbol\mu^{q-2}}{\norm{\boldsymbol\mu}_q^q}\Big)^{-1}
\end{eqnarray}
and
\begin{eqnarray}
\bar{\boldsymbol\chi}=\boldsymbol\chi\odot\Big(\frac{\boldsymbol\mu^{p-2}}{\norm{\boldsymbol\mu}_p^p}+\frac{\boldsymbol\mu^{q-2}}{\norm{\boldsymbol\mu}_q^q}\Big)^{-1}
\end{eqnarray}
As $\boldsymbol\mu\geq0$, $\mathbf{u}\geq0$ and $\boldsymbol\chi\geq0$, one has $\bar{\mathbf{u}}\geq0$, $\bar{\boldsymbol\chi}\geq0$. From Eq. \ref{sumeq} we have
\begin{eqnarray}
\boldsymbol\mu=(\bar{\mathbf{u}}+\bar{\boldsymbol\chi})/\gamma
\label{simp}
\end{eqnarray}
Due to \eqref{eq:subeq2} it must hold:
\begin{eqnarray}
\boldsymbol\chi^\top\boldsymbol\mu = \frac{1}{\gamma}\sum_{c,g}\chi_{c,g}(\bar{u}_{c,g}+\bar{\chi}_{c,g})=0
\label{zeroeq}
\end{eqnarray}
Note that in order for Eq. \ref{zeroeq} to be satisfied, it should hold that $\boldsymbol\chi=0$. In order to illustrate this, we employ contradiction and suppose all components of $\boldsymbol\chi$ are not zero and there exists an arbitrary index $j$ for which $\chi_j=\epsilon>0$. Such an assumption results in
\begin{eqnarray}
\boldsymbol\chi^\top\boldsymbol\mu=\frac{\epsilon}{\gamma}\Big(\frac{\mu_{j}^{p-2}}{\norm{\boldsymbol\mu}_p^p}+\frac{\mu_{j}^{q-2}}{\norm{\boldsymbol\mu}_q^q}\Big)^{-1}\Big({u}_j+\epsilon \Big)>0
\end{eqnarray}
which opposes the condition $\boldsymbol\chi^\top\boldsymbol\mu=0$. Hence, $\boldsymbol\chi$ does not include any non-zero components, and as a result, $\boldsymbol\chi=0$ which yields $\bar{\boldsymbol\chi}=0$. Based on Eq. \ref{simp}, $\boldsymbol\mu$ is determined as
\begin{eqnarray}
\boldsymbol\mu = \mathbf{\bar{u}}/\gamma
\end{eqnarray}
and due to $\norm{\boldsymbol\mu}_p\norm{\boldsymbol\mu}_q=1$, one obtains
\begin{equation}
\gamma = \sqrt{\norm{\bar{\mathbf{u}}}_p\norm{\bar{\mathbf{u}}}_q}
\label{gam}
\end{equation}

\section{Proof of Proposition 1}
\label{prop1proof}
As discussed in Section IV.D, thanks to the introduction of vector $\boldsymbol\mu$ as the collection of kernel weights from all clusters, the optimisation problem associated with the proposed approach given in Eq. (17) exactly matches that of \cite{9535490} given in Eq. (6). Consequently, following a similar optimisation approach, the convergence guarantees presented in \cite{9535490} directly carry over to the proposed localised approach. More specifically, for a sufficiently large $\delta$, Algorithm 1 is convergent to a single fixed-point independent of the initialisation of $\boldsymbol{\lambda}$.

\section{Proof of Proposition 2}
\label{prop2proof}
In order to compute $\mathcal{R}(H^{p,q}_{CG})$, let us first consider the following theorem from \cite{MohriRostamizadehTalwalkar18}
\begin{thm}[Rademacher complexity of a kernel-based hypothesis] Assume $\kappa :\mathcal{X} \times \mathcal{X}\rightarrow \mathbb{R}$
is a positive definite symmetric kernel function and suppose $\psi: \mathcal{X}\rightarrow \mathbb{H}$ is a feature projection linked to $\kappa$. If $S\subseteq\{
x:\kappa(x, x)\leq r^2\}$ is a dataset of size $n$, and $\mathbb{H}=\{x\rightarrow \langle\mathbf{w},\psi(x)\rangle:\norm{\mathbf{w}}_\mathbb{H}\leq \Lambda\}$
for some $\Lambda\geq0$ denotes the hypothesis class, then, the empirical Rademacher complexity of $\mathbb{H}$ over $S$ is bounded as
\begin{eqnarray}
\mathcal{R}(\mathbb{H})\leq\frac{\Lambda\sqrt{\mathrm{Tr}[\mathbf{K}]}}{n}
\end{eqnarray}
\noindent where $\mathrm{Tr}[\mathbf{K}]$ denotes the trace of the kernel matrix $\mathbf{K}$.
\label{KCR}
\end{thm}
\noindent Note that in the proposed approach, we have $\mathbf{K}=\sum_c\sum_g\mu_{cg}
\mathbf{K}_{cg}$, and hence
\begin{eqnarray}
\nonumber \mathrm{Tr}[\mathbf{K}]&=&\sum_i\sum_c\sum_g\mu_{cg}p^2_c(\mathbf{x}_i)\kappa_g(\mathbf{x}_i,\mathbf{x}_i)\\
\nonumber &\leq& r^2\sum_c\sum_g\mu_{cg}\sum_ip^2_c(\mathbf{x}_i)\\
&\leq&\Bigg(r^2\sum_c\sum_g\mu_{cg}\Bigg)\Bigg(\sum_c\sum_ip^2_c(\mathbf{x}_i)\Bigg)
\label{trr}
\end{eqnarray}
\noindent where the second inequality is correct as $\mu_{cg}$'s and $p_c(\mathbf{x}_i)$'s are non-negative. Note that $\sum_c\sum_g\mu_{cg}=\norm{\boldsymbol\mu}_1$. Using an $\ell_1$- to an $\ell_p$-norm ($p\geq1$) conversion \cite{JMLR:v12:kloft11a}, we have
\begin{eqnarray}
\norm{\boldsymbol\mu}_1\leq(CG)^{1-1/p}\norm{\boldsymbol\mu}_p
\end{eqnarray}
Similarly, for $q\geq1$ we have
\begin{eqnarray}
\norm{\boldsymbol\mu}_1\leq(CG)^{1-1/q}\norm{\boldsymbol\mu}_q
\end{eqnarray}
Multiplying the two inequalities above, one obtains
\begin{eqnarray}
\nonumber \norm{\boldsymbol\mu}_1^2&\leq& (CG)^{2-1/p-1/q}\norm{\boldsymbol\mu}_p\norm{\boldsymbol\mu}_q\\
&=&(CG)^{2-1/p-1/q}\norm{\boldsymbol\mu\boldsymbol\mu^\top}_{p,q}
\end{eqnarray}
Since in the proposed approach we have $\norm{\boldsymbol\mu\boldsymbol\mu^\top}_{p,q}\leq1$, one obtains
\begin{eqnarray}
\sum_c\sum_g\mu_{cg}=\norm{\boldsymbol\mu}_1\leq (CG)^{1-\frac{1}{2p}-\frac{1}{2q}}
\label{SPB}
\end{eqnarray}
Consequently, using Theorem \ref{KCR} and Eq. \ref{trr}, the empirical Rademacher complexity of the proposed approach is bounded as
 \begin{eqnarray}
 \mathcal{R}(H_{CG}^{p,q})\leq \frac{\Lambda r}{n}(CG)^{\frac{1}{2}-\frac{1}{4p}-\frac{1}{4q}}\Bigg(\sum_c\sum_ip_c^2(\mathbf{x}_i)\Bigg)^{1/2}
 \label{MRR}
 \end{eqnarray}

\section{Dependence of $\Lambda$ on $\delta$}
\label{depil}
Since $\boldsymbol\lambda^\top\mathbf{K}\boldsymbol\lambda\leq \Lambda^2$, and as the kernel matrix is constant given that the kernel function and the data are chosen and fixed, the upper bound on $\boldsymbol\lambda^\top\mathbf{K}\boldsymbol\lambda$ may be minimised by reducing the magnitude of $\boldsymbol\lambda$ which is controlled as
\begin{eqnarray}
\nonumber \norm{\boldsymbol\lambda}_2 &=& \norm{\Big(\delta \mathbf{I}+\mathbf{K}\Big)^{-1}\mathbf{1}}_2\leq \norm{\Big(\delta \mathbf{I}+\mathbf{K}\Big)^{-1}}_2\norm{\mathbf{1}}_2\\
&=&\frac{\sqrt{n}}{\vartheta+\delta}\leq \frac{\sqrt{n}}{\delta}=\frac{\theta}{\sqrt{n}}
\label{B1}
\end{eqnarray}
where the first equality is true since the norm of a vector-matrix product is less than or equal to their norms multiplied. In the equation above, $\vartheta$ stands for the least significant eigenvalue of the combined kernel matrix. Note that the eigenvalues are shifted by the amount of the constant added to the main diagonal and the norm of an inverse matrix is equal to the reciprocal of the minimal eigenvalue of the primary matrix. According to Eq. \ref{B1}, by increasing $\delta$, the magnitude of $\boldsymbol\lambda$ reduces, and consequently, the upper bound on $\boldsymbol\lambda^\top\mathbf{K}\boldsymbol\lambda$ may be reduced and subsequently, the empirical Rademacher complexity is decreased. However, shrinking the elements of $\boldsymbol\lambda$ may lead to an increase in the training error loss. As such, the optimal $\delta$ (or equivalently $\theta$) may be determined via cross validation.

\section{Proof of Proposition 4}
\label{prop3proof}
By inspecting Eq. \ref{trr}, it is clear that different regularisation constraints shall only lead to differences on the bounds obtained for the "$\sum_c\sum_g\mu_{cg}$" term.
\begin{enumerate}[label=(\roman*)]
\item For the joint vector $\ell_p$-norm regularisation ($\norm{\boldsymbol\mu}_p\leq1$), using an $\ell_1$-to-$\ell_p$ conversion we have
\begin{eqnarray}
\label{jointvector}
\nonumber \sum_c\sum_g\mu_{cg}=&\norm{\boldsymbol\mu}_1\leq(CG)^{1-1/p}\norm{\boldsymbol\mu}_p\mathrm{ }\\
&\overset{\norm{\boldsymbol\mu}_p\leq1}
\leq (CG)^{1-1/p}
\end{eqnarray}
\item For the disjoint vector $\ell_p$-norm regularisation ($\norm{\boldsymbol\mu_c}_p\leq1$, $c=1,\dots,C$) we have
\begin{eqnarray}
\nonumber \sum_c\sum_g\mu_{cg}&=&\sum_c\norm{\boldsymbol\mu_{c}}_1\leq \sum_c G^{1-1/p}\norm{\boldsymbol\mu_c}_p\\
&&\overset{\norm{\boldsymbol\mu_c}_p\leq1}\leq CG^{1-1/p}
\label{disjointvector}
\end{eqnarray}
\item And, for the disjoint matrix-norm regularisation ($\norm{\boldsymbol\mu_c\boldsymbol\mu_c^\top}_{p,q}\leq1$, $c=1,\dots,C$), using $\ell_1$-to-$\ell_p$ and $\ell_1$-to-$\ell_q$ conversions for $p,q\geq1$, we have
\begin{eqnarray}
\nonumber \norm{\boldsymbol\mu_c}_1\leq G^{1-1/p}\norm{\boldsymbol\mu_c}_p
\\\norm{\boldsymbol\mu_c}_1\leq G^{1-1/q}\norm{\boldsymbol\mu_c}_q
\end{eqnarray}
Multiplying the two inequalities above, one obtains
\begin{eqnarray}
\nonumber \norm{\boldsymbol\mu_c}_1^2&\leq& G^{2-\frac{1}{p}-\frac{1}{q}}\norm{\boldsymbol\mu_c}_p\norm{\boldsymbol\mu_c}_q\\
&=&G^{2-\frac{1}{p}-\frac{1}{q}}\norm{\boldsymbol\mu_c\boldsymbol\mu_c^\top}_{p,q}
\end{eqnarray}
Since $\norm{\boldsymbol\mu_c\boldsymbol\mu_c^\top}_{p,q}\leq1$, $\norm{\boldsymbol\mu_c}_1$ is bounded as
\begin{eqnarray}
\norm{\boldsymbol\mu_c}_1\leq G^{1-\frac{1}{2p}-\frac{1}{2q}}
\end{eqnarray}
Consequently, for the disjoint matrix-norm class we have
\begin{eqnarray}
\nonumber \sum_c\sum_g\mu_{cg}&=&\sum_c\norm{\boldsymbol\mu_{c}}_1\leq \sum_c G^{1-\frac{1}{2p}-\frac{1}{2q}}\\
&\leq& CG^{1-\frac{1}{2p}-\frac{1}{2q}}
\label{disjointmatrix}
\end{eqnarray}
\end{enumerate}
Comparing the bounds in Eq.'s \ref{jointvector}, \ref{disjointvector}, and \ref{disjointmatrix} with that of the proposed joint matrix-norm approach given in Eq. \ref{SPB}, the proposition is derived.$\square$

\vfill

\end{document}